\documentclass{article}

    \PassOptionsToPackage{numbers, compress}{natbib}

    \usepackage[preprint]{neurips_2025}

\usepackage[utf8]{inputenc} %
\usepackage[T1]{fontenc}    %
\usepackage{microtype,inconsolata}
\usepackage{times,latexsym}
\usepackage{graphicx} \graphicspath{{figures/}}
\usepackage{amsmath,amssymb,mathabx,mathtools,amsthm,nicefrac}
\usepackage[linesnumbered,ruled,vlined]{algorithm2e}
\usepackage{acronym}
\usepackage{enumitem}
\usepackage{balance}
\usepackage{xspace}
\usepackage{setspace}
\usepackage[skip=3pt,font=small]{subcaption}
\usepackage[skip=3pt,font=small]{caption}
\usepackage[dvipsnames,svgnames,x11names,table]{xcolor}
\usepackage{booktabs,tabularx,colortbl,multirow,multicol,array,makecell,tabularray}
\usepackage{overpic,wrapfig}
\usepackage[misc]{ifsym}
\usepackage{pifont}
\usepackage{tcolorbox}  %
\usepackage{listings}
\usepackage{mathrsfs}
\usepackage{upgreek}
\usepackage{vcell}
\usepackage{mwe} %
\usepackage{calc} %

\usepackage{epigraph}
\usepackage[pagebackref,breaklinks,colorlinks]{hyperref}
\usepackage[capitalise,noabbrev,nameinlink]{cleveref}

\usepackage[ruled,vlined]{algorithm2e}

\crefname{algorithm}{Alg.}{Algs.}
\Crefname{algocf}{Algorithm}{Algorithms}
\crefname{section}{Sec.}{Secs.}
\Crefname{section}{Section}{Sections}
\crefname{table}{Tab.}{Tabs.}
\Crefname{table}{Table}{Tables}
\crefname{figure}{Fig.}{Fig.}
\Crefname{figure}{Figure}{Figure}
\definecolor{revision}{RGB}{0,0,255}

\definecolor{iccvblue}{rgb}{0.21,0.49,0.74}

\definecolor{Gray}{gray}{0.85}

\newcommand{\method}{\texttt{\textbf{VLA-RL}}\xspace}
\newcommand{\methodfull}{VLA-RL\xspace}
\newcommand{\tech}{\mbox{{robotic process reward model}}\xspace}
\newcommand{\techsmall}{\mbox{{rprm}}\xspace}
\newcommand{\techLarge}{\mbox{{Robotic Process Reward Model}}\xspace}

\definecolor{best}{rgb}{0.96, 0.57, 0.58}
\definecolor{second}{rgb}{0.98, 0.78, 0.57}
\definecolor{third}{rgb}{1.0, 1.0, 0.56}

\definecolor{shallowgreen}{RGB}{200, 230, 201} %
\definecolor{shalloworange}{RGB}{255, 224, 178} %

\definecolor{citecolor}{HTML}{696FAD}
\hypersetup{
  colorlinks=true,
  citecolor=citecolor,
}

\acrodef{dit}[DiT]{Diffusion Transformer}
\acrodef{ddpm}[DDPM]{Denoising Diffusion Probabilistic Model}
\acrodef{vae}[VAE]{Variational Autoencoder}
\acrodef{il}[IL]{Imitation Learning}
\acrodef{rl}[RL]{Reinforcement Learning}
\acrodef{gcrl}[GCRL]{Goal-Conditioned Reinforcement Learning}
\acrodef{mbrl}[MBRL]{Model-based Reinforcement Learning}
\acrodef{pomdp}[POMDP]{Partially Observable Markov Decision Process}
\acrodef{mdp}[MDP]{Markov Decision Process}
\acrodef{3d}[3D]{three-dimensional}
\acrodef{nerf}[NeRF]{Neural Radiance Field}
\acrodef{sde}[SDE]{stochastic differential equation}
\acrodef{3dgs}[3D-GS]{3D Gaussian Splatting}
\acrodef{fps}[FPS]{Farthest Point Sampling}
\acrodef{dof}[DoF]{Degree of Freedom}
\acrodef{ood}[OOD]{Out-of-Domain}
\acrodef{ppo}[PPO]{Proximal Policy Optimization}
\acrodef{dpo}[DPO]{Direct Preference Optimization}
\acrodef{sft}[SFT]{Supervised Fine-tuning}
\acrodef{gae}[GAE]{Generalized Advantage Estimation}

\makeatletter
\DeclareRobustCommand\onedot{\futurelet\@let@token\@onedot}
\def\@onedot{\ifx\@let@token.\else.\null\fi\xspace}

\def\ie{\emph{i.e}\onedot}

\makeatother

\newlength\savewidth

\makeatletter
\renewcommand{\paragraph}{%
  \@startsection{paragraph}{4}{\z@}%
  {1ex plus 0.5ex minus 0.2ex} %
  {-1em}                      %
  {\normalfont\normalsize\bfseries} %
}
\makeatother

\usepackage{amsmath,amsfonts,mathtools,bm}

\def\eqref#1{equation~\ref{#1}}

\def\1{\bm{1}}

\DeclareMathAlphabet{\mathsfit}{\encodingdefault}{\sfdefault}{m}{sl}
\SetMathAlphabet{\mathsfit}{bold}{\encodingdefault}{\sfdefault}{bx}{n}

\title{\method: Towards Masterful and General Robotic Manipulation with Scalable Reinforcement Learning}

\author{
  Guanxing Lu$^{1}$, Wenkai Guo$^{2}$, Chubin Zhang$^{1}$, Yuheng Zhou$^{2}$, Haonan Jiang$^{1}$\\ \textbf{Zifeng Gao$^{1}$, Yansong Tang$^{1}$\setcounter{footnote}{1}\thanks{Corresponding author.}~, Ziwei Wang$^{2}$} \\
  $^1$ Tsinghua Shenzhen International Graduate School, Tsinghua University \\
  $^2$ School of Electrical and Electronic Engineering, Nanyang Technological University \\
  \href{https://github.com/GuanxingLu/vlarl}{\texttt{\textbf{github.com/GuanxingLu/vlarl}}}
}

\begin{document}

\maketitle

\begin{abstract}

Recent high-capacity vision-language-action (VLA) models have demonstrated impressive performance on a range of robotic manipulation tasks by imitating human demonstrations. 
However, exploiting offline data with limited visited states will cause execution failure in out-of-distribution scenarios.
Intuitively, an exploration-based method that improves on online collected data at test time could address this limitation. 
We present \method, an algorithmic and systematic framework that leverages online reinforcement learning (RL) to improve pretrained auto-regressive VLAs in downstream tasks.
Within a unified perspective, we first introduce a trajectory-level RL formulation for auto-regressive VLA training, which models general robotic manipulation trajectory as multi-modal multi-turn conversation. 
To address the challenge of sparse rewards, we fine-tune a pretrained vision-language model as a \tech, which is trained on pseudo reward labels annotated on automatically extracted task segments.
To scale up, we identify several implementation findings that improve the stability and efficiency including curriculum selection strategy, GPU-balanced vectorized environments, batch decoding, and critic warmup.
\method enables OpenVLA-7B to surpass the strongest finetuned baseline by $4.5\%$ on $40$ challenging robotic manipulation tasks in LIBERO, and even matches the performance of advanced commercial models such as $\pi_0$-FAST.
Notably, we observe that \method benefits from increased test-time optimization, indicating an early spark of inference scaling laws in robotics.

\end{abstract}

\epigraph{``We want AI agents that can discover like we can, not which contain what we have discovered.''}{--- \emph{The Bitter Lesson}}

\section{Introduction}
\label{sec:intro}

Large foundation models pretrained on internet-scale datasets have demonstrated effectiveness across a variety of domains, such as text~\cite{ouyang2022traininglanguagemodelsfollow,christiano2023deepreinforcementlearninghuman,deepseekai2025deepseekr1incentivizingreasoningcapability}, image~\cite{yang2024qwen2,wang2024ponder,ye2024voco,ye2024atp}, video~\cite{zhang2024flash,wang2024hierarchical}, and audio~\cite{chu2023qwen}.
Recently, large vision-language-action (VLA) models~\citep{brohan2022rt,brohan2023rt,octo_2023,kim2024openvla,lin2024datascalinglawsimitation,kalashnikov2018qt,mtopt2021arxiv,ehsani2023imitating,bharadhwaj2023roboagent,black2024pi_0,pertsch2025fast,intelligence2025pi_} have been proposed by imitating large-scale human demonstrations~\citep{padalkar2023open,fang2023rh20t,pinto2016supersizing,mandlekar2018roboturk,kalashnikov2018qt,gupta2018robot,dasari2019robonet,cabi2019,jang2022bc,ebert2021bridge,walke2023bridgedata,bharadhwaj2023roboagent,khazatsky2024droid}, which indicates a possible pathway towards generalist robots that can perform diverse manipulation tasks.
However, exploiting offline data with limited visited states will cause execution failure in \ac{ood} scenarios at test time.

We believe the key to overcoming such challenges lies in transforming \emph{exploitation-based} approaches into \emph{exploration-based} methods, as exemplified by \ac{rl}~\cite{silver2025welcome}.
Recent breakthroughs witnessed in applying RL to large language models (LLMs) have shown remarkable progress~\cite{deepseekai2025deepseekr1incentivizingreasoningcapability,
yu2025DAPO,liu2025DRGRPO,hu2025openreasonerzero}, where the scaling benefits from imitation pretraining on offline web data are approaching their limits. Consequently, reinforcement learning has emerged as a promising paradigm for achieving test-time scaling improvements by training on online collected data with unlimited state coverage.
It is natural to ask: \emph{can we achieve similar RL-based test-time scaling benefits in the field of robotic manipulation?}

On the other hand, many efforts have been proposed to apply RL to robotics \cite{tang2024deep}. However, traditional RL from scratch often suffers from data inefficiency and requires extensive reward engineering~\cite{akkaya2019solving,hu2023causal,zeng2024poliformer,zhu2020ingredients}.
Thus, previous works focus on simple domains, with low-dimensional state spaces~\cite{rajeswaran2017learning,gupta2019relay,nair2020awac} and small-scale network architecture (e.g. MLP)\cite{rajeswaran2017learning,uchendu2023jump,agarwal2022reincarnating,luo2024serl,luo2024hilserl}, single-task learning \cite{zhu2018reinforcement,kober2010imitation}. 
In contrast, fine-tuning from large robotics foundation models with rich representational knowledge may significantly reduce the search space and enable the model to learn complex motion patterns, making training on general tasks and environments possible.

We explore this question through a systematic study.
To efficiently implement scalable \ac{rl} training, we present \method, a unified framework that leverages online \ac{rl} to improve pretrained auto-regressive VLAs.
Specifically, within a unified perspective, we first introduce a general \ac{rl} formulation for auto-regressive VLA training, which models general robotic manipulation trajectory as multi-modal multi-turn conversation. 
To address challenges associated with sparse rewards in the expansive robot action space, we instantiate a \tech as vision-language model finetuned on automatically extracted pseudo reward labels.
Based on \method, we identify systematic implementation improvements, including task selection, GPU-balanced environments, batch decoding, and critic warmup—to enhance stability and efficiency.
Empirically, we adopt OpenVLA-7B~\cite{kim2024openvla} as the base VLA and apply our method to $40$ challenging robotic tasks in LIBERO~\cite{liu2024libero}.
The results show that \method improves the base model by a large margin of $4.5$\% and even matches the performance of advanced commercial models such as $\pi_0$-FAST~\cite{pertsch2025fast}.
Furthermore, \method's performance consistently improves with increased test-time computation, suggesting preliminary evidence of inference scaling laws~\cite{snell2024scaling} in robotics.

\begin{figure}[t]
    \centering
    \begin{subfigure}{0.65\textwidth}
        \centering
        \includegraphics[width=\textwidth]{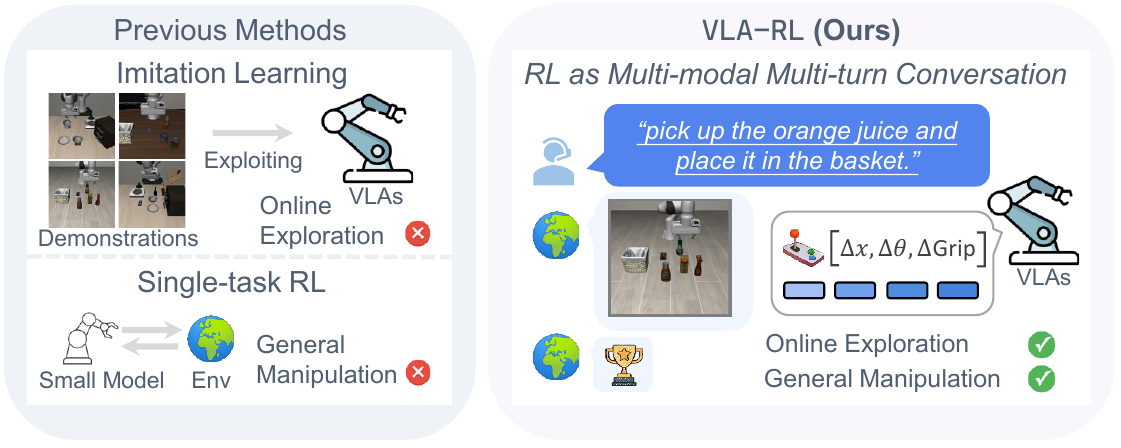}
    \end{subfigure}
    \hfill
    \begin{subfigure}{0.33\textwidth}
        \centering
        \includegraphics[width=\textwidth]{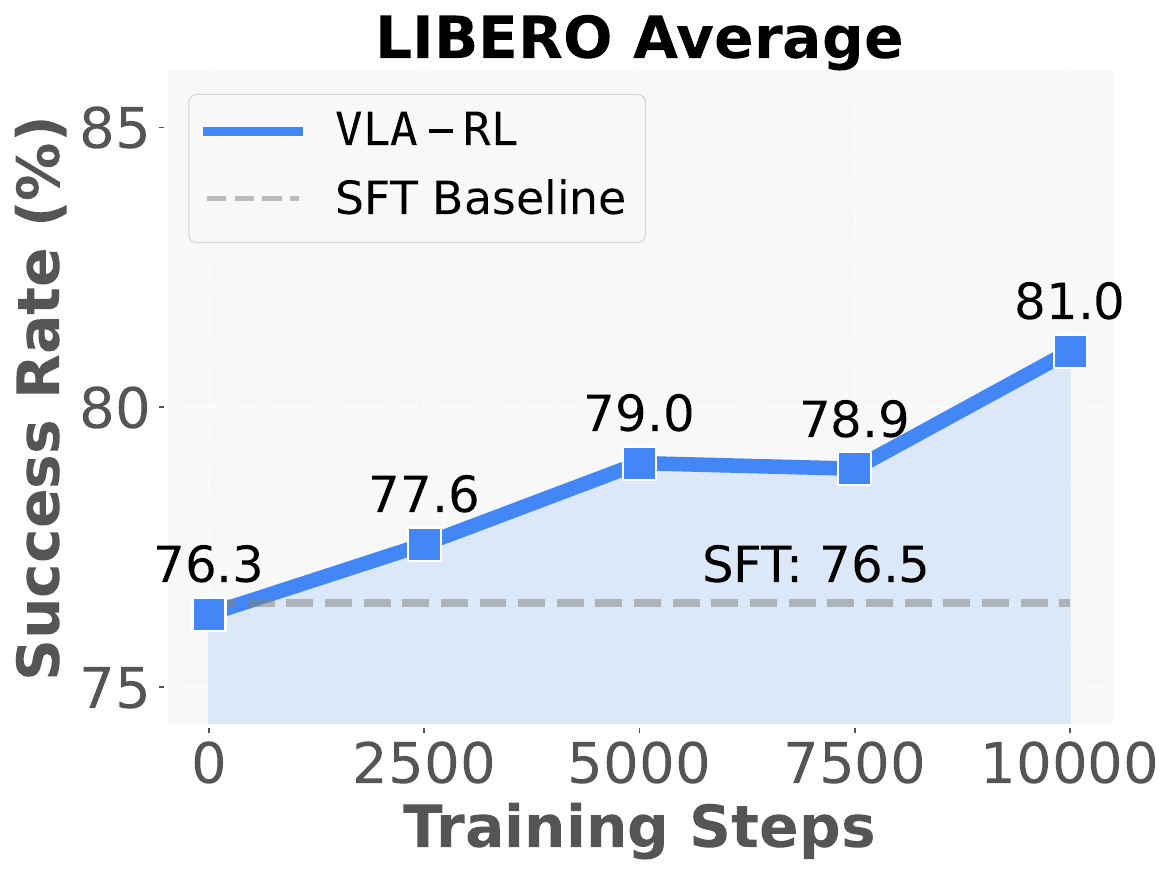}
    \end{subfigure}
    \caption{\small Previous VLAs focus on imitation learning that exploits the offline demonstrations, while \method explores improving high-capacity VLAs with scalable reinforcement learning. For evaluation, we train OpenVLA-7B to master $40$ challenging robotic manipulation tasks in LIBERO, and show a notable consistent improvement over the imitation learning baseline.}
    \label{fig:teaser}
\end{figure}

\section{Related Work}
\label{sec:related_work}

\noindent\textbf{Robotic Foundation Models.} 
Robotic foundation models demonstrating remarkable potential in developing generalizable robot behaviors through multi-task training~\citep{brohan2022rt,brohan2023rt,octo_2023,kim2024openvla,lin2024datascalinglawsimitation,kalashnikov2018qt,mtopt2021arxiv,ehsani2023imitating,bharadhwaj2023roboagent,black2024pi_0,pertsch2025fast,intelligence2025pi_,lu2023thinkbot,liu2024plan} on extensive multi-task and multi-embodiment robot datasets~\citep{padalkar2023open,fang2023rh20t,dasari2019robonet,ebert2021bridge,walke2023bridgedata,khazatsky2024droid}. 
Notably, OpenVLA-7B~\cite{kim2024openvla} finetunes high-capacity vision-language models to generate robot actions as language tokens in the model vocabulary, depicting impressive generalizability across various tasks. 
However, these VLAs trained through imitation learning face significant deployment challenges when dealing with \ac{ood} scenarios that are not covered in the offline expert demonstrations.

\noindent\textbf{Reinforcement Learning for Robotics Models.} 
Although reinforcement learning has demonstrated numerous achievements in robotics~\cite{tang2024deep}, implementing RL methods from the ground up typically demands sophisticated reward and training paradigm designs~\cite{akkaya2019solving,hu2023causal,zeng2024poliformer,zhu2020ingredients,hu2024flare,zhang2024grape,guo2025improving}.
Consequently, researchers have thoroughly investigated the utilization of pretrained models to enhance RL performance \cite{taiga2023investigating,khetarpal2022towards,taylor2009transfer,nair2020awac,hester2018deep,gupta2019relay,julian2020never,lu2021aw,baker2022video,zhu2023transfer,wołczyk2024finetuningreinforcementlearningmodels,uchendu2023jump, hu2023imitation,xing2024bootstrapping}.
However, prior studies necessitate complete offline dataset availability throughout the fine-tuning process\cite{nair2020awac,agarwal2022reincarnating,hester2018deep,gupta2019relay,julian2020never,vecerik2017leveraging,kober2010imitation,lu2021aw,rajeswaran2017learning,ball2023rlpd}. 
Furthermore, other proposed methodologies are exclusively tested in environments with simplified state representations~\cite{rajeswaran2017learning,gupta2019relay,nair2020awac}, naive neural network structures~\cite{rajeswaran2017learning,uchendu2023jump,agarwal2022reincarnating,luo2024serl,luo2024hilserl}, and narrow single-task training paradigms~\cite{zhu2018reinforcement,kober2010imitation}.
In contrast, \method explores fine-tuning from large robotics foundation models with trajectory-level \ac{rl}, where the rich representational knowledge significantly reduces the search space and enables the model to learn complex motion patterns, making training on general tasks and environments possible.

\noindent\textbf{Reinforcement Learning for Large Models.} 
Reinforcement learning has significantly enhanced LLMs' reasoning capabilities through approaches that inspire our work~\cite{schulman2017proximal,deepseekai2025deepseekr1incentivizingreasoningcapability,zhang2025lessonsdevelopingprocessreward,lightman2023verify,ouyang2022traininglanguagemodelsfollow,chen2021decisiontransformer,wang2025ragen,yu2025DAPO, liu2025DRGRPO,hu2025openreasonerzero,bai2025univg}.
This family of methods begins with STaR~\citep{zelikman2022star} and includes reinforced self-training~\citep{gulcehre2023reinforced} and rejection fine-tuning~\citep{yuan2023scaling}, relying on solutions generated by LLMs for interactive self-updating.
To facilitate more autonomous exploration and exploitation, policy optimization methods like PPO \citep{schulman2017proximal} are widely used in current practice. For example, GRPO \citep{deepseekai2025deepseekr1incentivizingreasoningcapability} strengthens systematic problem-solving capabilities particularly relevant to sequential decision-making. 
Approaches enabling reasoning include Process Reward Models \citep{zhang2025lessonsdevelopingprocessreward, lightman2023verify} that evaluate intermediate reasoning, and conversation-based training methods \citep{ouyang2022traininglanguagemodelsfollow,chen2021decisiontransformer,wang2025ragen} that optimize multi-turn interactions. 
Recent proposed RL techniques~ \citep{yu2025DAPO, liu2025DRGRPO,hu2025openreasonerzero} demonstrate that different implementation details significantly improve the scaling of reasoning in language models. Our work bridges these advances to robotics by formulating trajectory-level optimization as a multi-modal, multi-turn conversation, enabling robots to benefit from similar RL frameworks that have proven effective for large language models.

\begin{figure*}[t]
    \centering
    \includegraphics[width=1\textwidth]{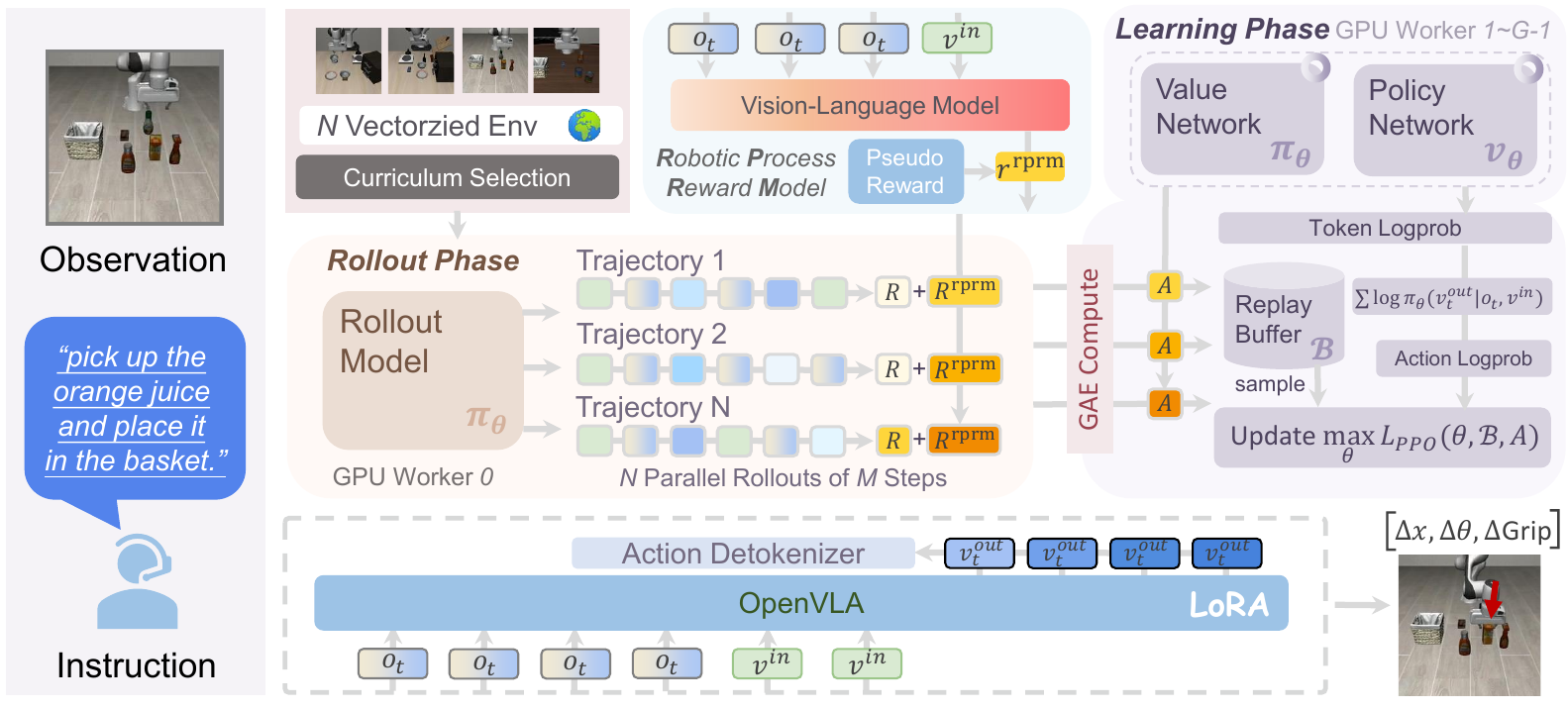}
    \caption{\small The overall pipeline of \method, which is composed of a transformer-based policy, a homogeneous value model, a frozen robotic process reward model, and the vectorized environments.}
    \label{fig:pipeline}
\end{figure*}

\section{\methodfull}
\label{sec:approach}

In this section, we start with brief preliminaries on the problem descriptions and open-sourced VLA models (\cref{subsec:prelimilaries}). We present an overview of our pipeline (\cref{subsec:overall}). Subsequently, we introduce the mathematical formulation of general robotic manipulation as multi-modal multi-turn conversation (\cref{subsec:training}). Then we present \tech for reward densification (\cref{subsec:reward}). To implement in practice, we build and present the \method system (\cref{subsec:system_techniques}) with several key findings that facilitate scalable VLA training with \ac{rl}. 

\begin{algorithm*}[t]
\caption{\method}
\DontPrintSemicolon
\label{alg:vlarl}

\KwIn{number of environments $N$, initial VLA policy with parameters $\theta_0$, trained \tech with parameters $\phi$, post-processing function $f$, steps per update $M$, discount factor $\gamma$, GAE parameter $\lambda_{\text{GAE}}$}
\KwOut{Improved VLA policy $\theta_K$}

$\text{envs} \leftarrow \text{VecEnv}(\text{num\_envs}=N)$ \\
$\mathbf{o}_0 \leftarrow \text{envs.reset()}$ \\
$\mathbf{d}_0 \leftarrow \{0\}^N$ {\color{Gray} \quad \# Terminal state indicators} \\

\For{$k=0$ \KwTo $K-1$}{
  $\mathcal{B}_k \leftarrow \emptyset$ \\

  {\color{Gray} \# Rollout Phase: Collect trajectories from $N$ parallel environments} \\
  \For{$t=0$ \KwTo $M-1$}{
    $\mathbf{v}_t^{\text{in}} \leftarrow h(\mathbf{o}_t)$ {\color{Gray} \quad \# Apply vision-text processor to all observations} \\
    $\mathbf{v}_t^{\text{out}} \leftarrow \pi_{\theta_k}(\mathbf{o}_t, \mathbf{v}_t^{\text{in}})$ {\color{Gray} \quad \# Policy model generates} \\
    $\mathbf{a}_t \leftarrow f(\mathbf{v}_t^{\text{out}})$ {\color{Gray} \quad \# Extract actions} \\

    {\color{Gray} \# Compute action log probabilities as sum of token log probs} \\
    $\log \boldsymbol{\pi}_{\theta_k}(\mathbf{a}_t|\mathbf{o}_t,\mathbf{v}_t^{\text{in}}) \leftarrow \sum_i \log \boldsymbol{\pi}_{\theta_k}(\mathbf{v}^{\text{out}}_{t,i}|\mathbf{o}_t,\mathbf{v}_t^{\text{in}})$ \\ %

    $\mathbf{V}_t \leftarrow V_{\theta_k}(\mathbf{o}_t, \mathbf{v}_t^{\text{in}})$ {\color{Gray} \quad \# Value model estimates} \\
    $\mathbf{o}_{t+1}, \mathbf{r}_t^{\text{sparse}}, \mathbf{d}_{t+1}, \mathbf{i}_t \leftarrow \text{envs.step}(\mathbf{a}_t)$ \\
    $\mathbf{r}_t^{\text{\techsmall}} \leftarrow R_{\phi}(\mathbf{o}_t, \mathbf{a}_t)$ {\color{Gray} \quad \# Robotic process reward model} \\
    $\mathbf{r}_t \leftarrow \mathbf{r}_t^{\text{sparse}} + \mathbf{r}_t^{\text{\techsmall}}$ \\

    $\mathcal{B}_k \leftarrow \mathcal{B}_k \cup \{(\mathbf{o}_t, \mathbf{a}_t, \mathbf{r}_t, \mathbf{d}_t, \mathbf{v}_t^{\text{out}}, \log \boldsymbol{\pi}_{\theta_k}(\mathbf{a}_t|\mathbf{o}_t, \mathbf{v}_t^{\text{in}}), \mathbf{V}_t)\}$ \\
  }

  $\mathbf{A}^{\text{GAE}} \leftarrow \text{GAE}(\mathcal{B}_k, \mathbf{V}_M, \gamma, \lambda_{\text{GAE}})$ {\color{Gray} \# Compute advantages} \\

  $\theta_{k+1} \leftarrow \text{PPO}(\theta_k, \mathcal{B}_k, \mathbf{A}^{\text{GAE}})$ {\color{Gray} \quad \# Learning Phase: Update policy parameters} \\
}

\Return $\theta_K$
\end{algorithm*}

\subsection{Preliminaries}\label{subsec:prelimilaries}

General robotic manipulation has been a core pursuit of the robotic community for a long time. The agent is required to interactively determine the next robot action (the end-effector's pose) to perform diverse tasks, based on the visual observation and the human instruction that specifies the current task. Recently, high-capacity, pretrained vision-and-language models have demonstrated generalizability across a wide range of language-conditioned manipulation tasks.
Among these, OpenVLA-7B~\cite{kim2024openvla} is a leading open-source VLA, which hence acts as the base model of our method. 
At its core lies an auto-regressive LLM Llama-2-7B~\citep{touvron2023llama2} with a two-stream visual encoder that consists of pretrained SigLIP~\citep{zhai2023sigmoid} and DinoV2~\citep{oquab2023dinov2} models.
At each timestep $t$, it takes the captured image $\mathbf{o}_t$ by a third-person camera and the human instruction $\mathbf{v}_t^{\text{in}}$ as input and outputs an action token sequence $\mathbf{v}_t^{\text{out}} \in \mathcal{V}^n$, where each action token represents a discrete bin of one dimension of the robot action space. The final robot action is extracted from this sequence using a post-processing function $f$, resulting in $\mathbf{a}_t = f(\mathbf{v}_t^{\text{out}})$. 
However, optimizing the auto-regressive VLA poses challenges in both \emph{algorithmic} and \emph{systematic} aspects, including \ac{rl} for general manipulation, sparse reward issue, and large-scale evaluation and optimization, etc.

\subsection{Overall Pipeline}\label{subsec:overall}

The overall pipeline of \method is shown in \cref{fig:pipeline}, in which we develop an algorithmic and systematic framework to train auto-regressive VLAs with \ac{rl}. 
The system consists of three models, including the policy and value model that need to be trained in the commonly used actor-critic framework, and a frozen \tech that densifies the sparse reward given by the environment.
At the algorithmic level, we formulate auto-regressive \method training as a multi-modal and multi-turn conversation.
The systematic techniques such as GPU-balanced vectorized environment, batch decoding, curriculum selection strategy, and critic warmup further contribute to the training efficiency and stability of the system.
Finally, the post-trained VLA model is able to produce feasible actions by optimizing the expected reward, thereby performing diverse manipulation tasks successfully.

\subsection{General Robotic Manipulation as Multi-turn Conversation}\label{subsec:training}
To extend \ac{rl} to optimize auto-regressive VLAs for general manipulation, we first formulate the Markovian Decision Process as multi-turn conversation.
Let $\mathcal{V}$ represent the discrete, finite set of vocabulary tokens. The spaces $\mathcal{V}^m$ and $\mathcal{V}^n$ denote the possible input and output text sequences, with $m$ and $n$ specifying the maximum sequence lengths for inputs and outputs, respectively. We define the state space as the Cartesian product $\mathcal{S} = \mathcal{O} \times \mathcal{V}^m$, where $\mathcal{O}$ is the image space. The action space is given by the set of all possible output utterances, $\mathcal{V}^n$, generated by VLAs.
Accordingly, the VLA policy parameterized by $\theta$ can be formalized as a mapping $\pi_\theta: \mathcal{O} \times \mathcal{V}^m \to \mathcal{V}^n$. At each timestep, the policy assigns a probability $\pi_\theta(\mathbf{v}_t^{\text{out}} \mid o_t, \mathbf{v}_t^{\text{in}}) \in [0, 1]$ to emitting the output sequence $\mathbf{v}_t^{\text{out}}$ given the input image $o_t$ and prompt $\mathbf{v}_t^{\text{in}}$.
The environment’s transitions are governed by the function $\mathcal{T}: \mathcal{S} \times \mathcal{A} \mapsto \mathcal{S}$, describing how states evolve after each action.
The environmental reward function, $R: \mathcal{S} \to \mathbb{R}$, quantifies the quality of outcomes as $r_t$ following each action taken. 
Over a trajectory of $T$ timesteps, the objective is to maximize the discounted sum of rewards, $R^{\gamma} = \sum{t=0}^T \gamma^{t} r_t$, where $\gamma$ is the discount factor. We employ \ac{ppo} \citep{schulman2017proximal} for stable policy optimization.

\noindent\textbf{Rollout Phase.} We first merge the updated LoRA~\cite{hu2022lora} weights with the original checkpoint and broadcast it to the inference engine. Then the agent interacts with the environment according to its current policy $\pi_{\theta_{old}}$, generating a sequence of states, actions, and rewards (\ie, trajectories). The log-probability of an action sequence can be decomposed into the summation of token-level log probabilities in auto-regressive models:
\begin{equation}
\log \boldsymbol{\pi}_{\theta}(\mathbf{a}_t|\mathbf{o}_t,\mathbf{v}_t^{\text{in}}) = \sum_{i=1}^{|\mathcal{A}|} \log \boldsymbol{\pi}_{\theta}(\mathbf{v}^{\text{out}}_{t,i}|\mathbf{o}_t,\mathbf{v}_t^{\text{in}})
\end{equation}
where $|\mathcal{A}|=7$ is the degrees of freedom of the action space of OpenVLA.

\noindent\textbf{Learning Phase.} The PPO objective function utilizes importance sampling with clipping to ensure stable updates:
\begin{equation}
\mathcal{L}_{\text{ppo}}(\theta) = \mathbb{E}_t\left[\min\left(\frac{\pi_\theta(\mathbf{a}_t|\mathbf{o}_t,\mathbf{v}_t^{\text{in}})}{\pi_{\theta_{old}}(\mathbf{a}_t|\mathbf{o}_t,\mathbf{v}_t^{\text{in}})}A_t, \text{clip}\left(\frac{\pi_\theta(\mathbf{a}_t|\mathbf{o}_t,\mathbf{v}_t^{\text{in}})}{\pi_{\theta_{old}}(\mathbf{a}_t|\mathbf{o}_t,\mathbf{v}_t^{\text{in}})}, 1-\epsilon, 1+\epsilon\right)A_t\right)\right]
\end{equation}
where $\epsilon$ is the clipping parameter that restricts the ratio between the new and old policies, preventing excessive policy updates.  For each state, the advantage $A_t$ is computed via \ac{gae}~\cite{schulman2015high}. The overall process is summarized in \cref{alg:vlarl}.

\subsection{\techLarge}\label{subsec:reward}

Reward modeling is the crux of applying \ac{rl} to general manipulation, which needs to: (1) provide dense rewards in environments with naturally sparse feedback; (2) avoid reward hacking where agents exploit the reward function in unintended ways. To this end, we propose \tech, a novel approach for reward densification that aligns with the VLA's token generation process.

\noindent\textbf{Reward Modeling as Next-token Prediction.} 
Traditional reinforcement learning in robotics often suffers from sparse rewards, typically binary signals provided only upon task completion. We reformulate reward modeling as a next-token prediction problem, leveraging the auto-regressive nature of pretrained vision-language models. Given a trajectory of states and actions, Robotic Process Reward Model (RPRM) predicts the likelihood of successful action sequences. The training objective is to maximize the log-likelihood of promising action tokens, weighted by a pseudo-reward signal that indicates progress towards task completion:
\begin{equation} \label{eq:prm_objective}
\mathcal{L}_{\text{rprm}}(\phi) = - \mathbb{E}_{t} \left[ \sum_{j=1} \log p_{\phi}(\mathbf{v}^{\text{rprm}}_{t,j} | \mathbf{v}^{\text{out}}_{t,<j}, \mathbf{o}_{t}, \mathbf{v}_{t}^{\text{in}}) \right]
\end{equation}
where $\phi$ are the parameters of the \tech model, $p_{\phi}(\mathbf{v}^{\text{rprm}}_j | \cdot)$ is the predicted probability of the next token $\mathbf{v}^{\text{rprm}}_j$ by \tech.

\noindent\textbf{Autonomous Pseudo Reward Label Generation.}
To train the \tech effectively without extensive human labeling, we develop an autonomous label generation pipeline that creates high-quality pseudo-reward labels from successful trajectories: (1) Milestone Segmentation: We collect a dataset of diverse successful trajectories from expert demonstrations and previous model runs. We segment trajectories into subtasks based on significant changes in gripper openness, as these often signify the completion of a functional step. (2) Progress Labeling: Within each segmented subtask, we identify keyframes where the robot's end-effector velocity approaches zero. These points often correspond to stable states or the completion of fine-grained motions. A positive pseudo-reward is assigned to the VLA action sequences leading to these keyframes.

The final reward is the direct summation of the golden sparse reward and the predicted reward from \tech. Our empirical analysis shows that this approach substantially accelerates learning while maintaining strong correlation with actual task success.

\subsection{The \method System}\label{subsec:system_techniques}

As \ac{rl} performance highly depends on the implementation details, we would like to share some tricks we have adopted in this project to increase the learning efficiency and stability.

\noindent\textbf{Curriculum Selection Strategy.}
We implement an adaptive curriculum that selects tasks based on the agent's current capabilities. For each task consists of an instruction and an initial state, we track success rates $s_j$ and compute sampling probabilities as:
\begin{equation}\label{eq:curriculum}
P(task_j) \propto \exp\left((0.5 - s_j)/\tau\right)
\end{equation}
where $\tau$ controls exploration. This equation prioritizes tasks with $\sim50\%$ success rate as the frontier of the agent's capabilities, while maintaining exposure to both mastered and challenging tasks, improving sample efficiency and generalization.

\noindent\textbf{Critic Warmup.} When training the value model (critic) from scratch, it initially produces inaccurate value estimates, which can mislead the model during the early stage of training.
To address this issue, we implement a critic warmup phase where we collect initial trajectories using the 
imitation pretrained policy and train the value network exclusively for several iterations before joint policy-value optimization begins.

\noindent\textbf{GPU-balanced Vectorized Environments}
We implement multiple vectorized environments for parallel rollout, where each training GPU holds a subset of environments.
Modern renderers often rely on GPUs for acceleration, but as the number of vectorized environments increases, GPU memory consumption can grow significantly. To address this, we assign each GPU worker its own set of environments to interact with and learn from. At the same time, we use an `all\_reduce` operation to gather the environmental states across all workers for the inference engine.

\noindent\textbf{Infrastructure.}
The PPO infrastructure uses bfloat16 to fit models into memory. Given $G$ GPUs in total, we allocate a dedicated $1$ GPU to do inference with vLLM~\cite{kwon2023efficient} acceleration and other $G\!-\!1$ GPUs for learning by Ray~\cite{moritz2018ray}, like done in OpenRLHF~\cite{hu2024openrlhf} and open-instruct~\cite{lambert2024tulu}. 
In our codebase, we have implemented OpenVLA~\cite{kim2024openvla} in the vLLM plugins to avoid using the original Huggingface transformers generation function, which leads to wrong results when dealing with large batch size.
The distributed training process is managed by PyTorch Fully Sharded Data Parallel (FSDP)~\cite{zhao2023pytorch} to support large-scale training.

\section{Experiments}
\label{sec:experiments}

In our experiments, we focus on the following questions:
\begin{enumerate}
    \item How does \method perform in commonly-used robotic manipulation benchmarks?
    \item Does \method scale with test time computing in general robotic manipulation?
    \item Why does \method show higher robustness than behavior cloning in terms of the state and action coverage of the training data?
    \item How do the proposed techniques and implementation details impact \method's performance? 
\end{enumerate}
In the following sections, we describe these key topics in detail with carefully controlled experiments.

\begin{figure*}[t]
    \centering
    \includegraphics[width=1\textwidth]{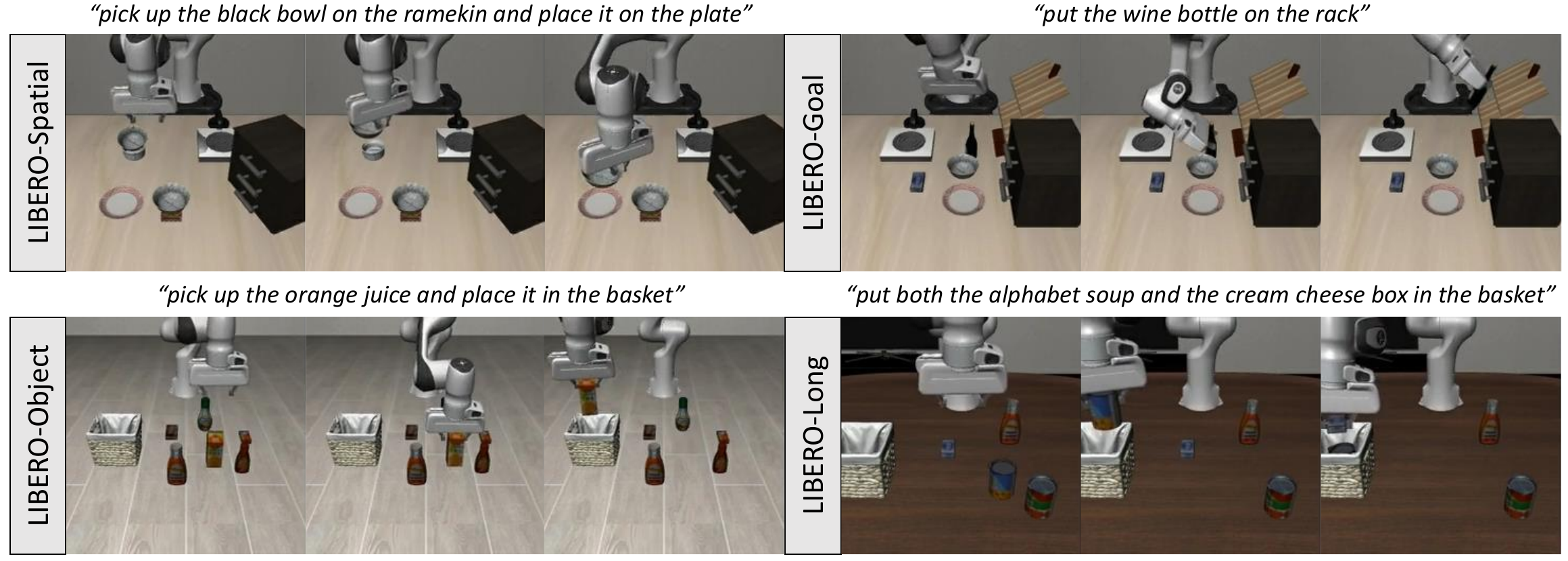}
    \caption{\small \textbf{Environments and Tasks.} For simulation, we evaluate on a commonly-used robotic manipulation benchmark named LIBERO with four task suites that focus on different challenges.
    }
    \label{fig:exp_setup}
\end{figure*}

\subsection{Major Experiments}
\label{subsec:quantitative_results}

\noindent\textbf{Experimental Setup.}
This section explores adapting \method to \emph{simulated} robot setups and tasks, specifically utilizing the \textbf{LIBERO} benchmark \citep{liu2024libero}.
The LIBERO benchmark~\citep{liu2024libero} encompasses four challenging task suites: LIBERO-Spatial, LIBERO-Object, LIBERO-Goal, and LIBERO-Long (or LIBERO-10), focusing on various spatial relationships, object categories, goal objectives, and extended sequential challenges. In our experiments, we focus on conducting \ac{rl} starting from a base model OpenVLA-7B~\cite{kim2024openvla} trained with supervised fine-tuning (SFT) for each task suite. In the test phase, all counterparts are evaluated across $500$ episodes for each suite. \cref{fig:exp_setup} shows several successful samples of the benchmark.

\noindent\textbf{Baselines.} 
For \ac{rl} training, we use the released checkpoints from \cite{kim2024openvla} as the base SFT model.
Besides the SFT baseline, we report the performance of Diffusion Policy~\citep{chi2023diffusionpolicy} trained from scratch, fine-tuned diffusion-based VLA Octo~\citep{octo_2023}, and GRAPE~\cite{zhang2024grape} trained with \ac{dpo}~\cite{rafailov2023direct} for better reference. In terms of metrics, we report the average success rates (SR) and the average ranking following \cite{kim2024openvla}.

\noindent\textbf{Performance Comparisons.}
We present the LIBERO experimental results in \cref{tab:libero_sim_results_table} after the \ac{rl} training process. 
\method improves the OpenVLA-7B checkpoint with SFT and \ac{dpo} by sizable margins of $4.5\%$ and $1.8\%$ respectively, which demonstrates the effectiveness of the online \ac{rl} and the proposed \method framework.
Notably, after only $48$ GPU hours of \ac{rl} training, the fine-tuned OpenVLA-7B matches the performance of an advanced commercial model $\pi_0$-FAST trained with high-quality SFT data, while the increasing trends are still consistent. The results imply the unlimited potential of boosting high-capacity VLAs with large-scale \ac{rl}.

\begin{figure*}[t]
    \centering
    \includegraphics[width=1\textwidth]{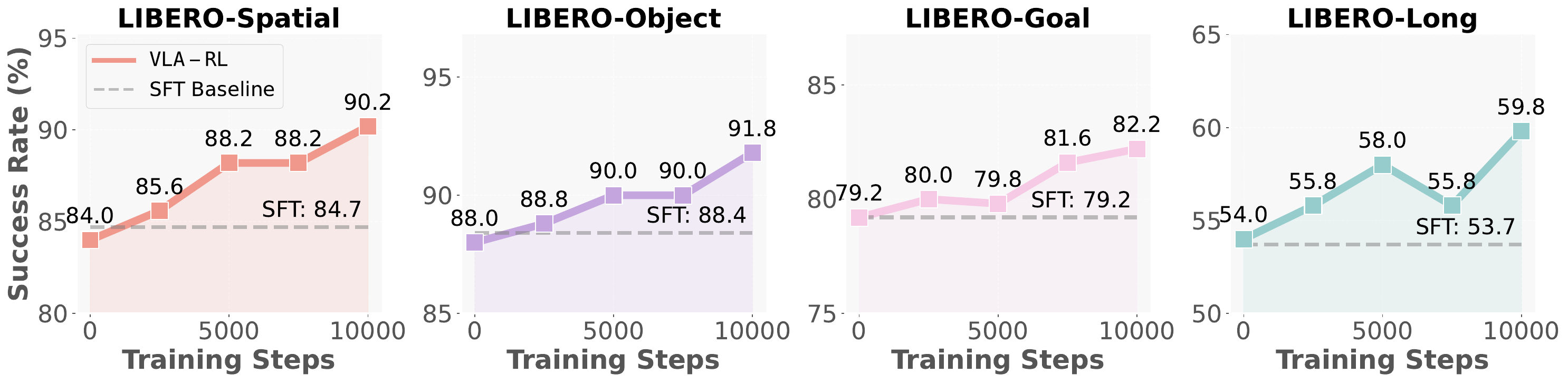}
    \caption{\small \textbf{Test-time Scaling Curve.}  We evaluate the fine-tuned OpenVLA-7B every $2500$ training steps on the complete suite and report the average task success rates.
    }
    \label{fig:test_time_scaling_curve}
\end{figure*}

\noindent\textbf{Test-time Scaling.}
We report the test success rates along the \ac{rl} training process of four LIBERO tasks in \cref{fig:test_time_scaling_curve}. The evaluation success rates on all four task suites are consistently improved with the test time optimization, indicating \emph{an early spark of inference scaling laws~\cite{snell2024scaling} in robotics}.

\begin{table*}[t]
\centering
\caption{\textbf{LIBERO simulation benchmark results.} We present average success rates and ranks based on 500 evaluation episodes for each method across LIBERO's four task suites. Methods are ranked $1\sim5$ within each suite, with 1 indicating best performance and 5 indicating worst. We also add an advanced commercial auto-regressive model {\color{gray}$\pi_0$-FAST} \cite{pertsch2025fast} for reference.}
\label{tab:libero_sim_results_table}
\resizebox{\textwidth}{!}{\begin{tabular}{l|cc|cc|cc|cc|cc}
\toprule
& \multicolumn{2}{c|}{LIBERO-Spatial} & \multicolumn{2}{c|}{LIBERO-Object} & \multicolumn{2}{c|}{LIBERO-Goal} & \multicolumn{2}{c|}{LIBERO-Long} & \multicolumn{2}{c}{Average} \\
\cline{2-3} \cline{4-5} \cline{6-7} \cline{8-9} \cline{10-11}
\addlinespace[2pt]
& SR ($\uparrow$) & Rank ($\downarrow$)& SR ($\uparrow$) & Rank ($\downarrow$)& SR ($\uparrow$) & Rank ($\downarrow$)& SR ($\uparrow$) & Rank ($\downarrow$)& SR ($\uparrow$) & Rank ($\downarrow$)\\
\midrule
Diffusion Policy \cite{chi2023diffusionpolicy} & 78.3\% & 5 & \textbf{92.5\%} & \textbf{1} & 68.3\% & 5 & 50.5\% & 5 & 72.4\% & 4.0 \\
Octo (SFT) \cite{octo_2023} & 78.9\% & 4 & 85.7\% & 5 & \textbf{84.6\%} & \textbf{1} & 51.1\% & 4 & 75.1\% & 3.5 \\
OpenVLA (SFT) \cite{kim2024openvla} & 84.7\% & 3 & 88.4\% & 4 & 79.2\% & 4 & 53.7\% & 3 & 76.5\% & 3.5 \\
GRAPE (DPO) \cite{zhang2024grape} & 87.6\% & 2 & 91.2\% & 3 & 82.2\% & 2 & 55.8\% & 2 & 79.2\% & 2.3 \\
{\color{gray}$\pi_0$-FAST} \cite{pertsch2025fast} & {\color{gray}96.4\%} & {\color{gray}-} & {\color{gray}96.8\%} & {\color{gray}-} & {\color{gray}88.6\%} & {\color{gray}-} & {\color{gray}60.2\%} & {\color{gray}-} & {\color{gray}85.5\%} & {\color{gray}-} \\
\midrule
\textbf{\method (Ours)} & \textbf{90.2\%} & \textbf{1} & 91.8\% & 2 & 82.2\% & 2 & \textbf{59.8\%} & \textbf{1} & \textbf{81.0\%} & \textbf{1.5} \\
\bottomrule
\end{tabular}}
\end{table*}

\begin{figure*}[t]
    \centering
    \includegraphics[width=1\textwidth]{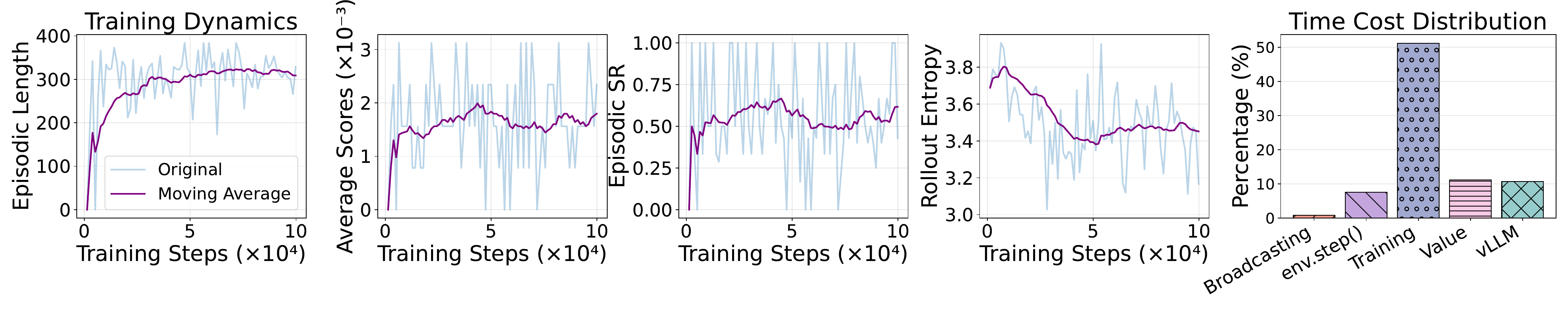}
    \caption{\small \textbf{Training Dynamics.} We draw the length of generated episodes, reward dynamics and rollout entropy along the training process on LIBERO-Long.
    }
    \vspace{-0.4cm}
    \label{fig:training_dynamics}
\end{figure*}

\subsection{Training Dynamics.}\label{subsec:training_dynamics}

Reinforcement learning on large models requires careful monitoring of key metrics to identify discrepancies and refine the system~\cite{yu2025DAPO}. Following the common practice~\cite{yu2025DAPO, lambert2024tulu}, we report and analyze the learning dynamics to share some key findings regarding training VLAs with \ac{rl}. The results are shown in \cref{fig:training_dynamics}.

\noindent\textbf{Length of Generated Episodes}
We observe that successful \method training leads to gradually decreasing episode lengths, indicating the model learns more efficient action sequences for manipulation tasks. This contrasts with LLM-based \ac{rl} where longer sequences often correlate with improved reasoning capability.

\noindent\textbf{Dynamics of Reward During Training.}
Reward trends show consistent improvement across training, with periodic plateaus corresponding to curriculum transitions between task difficulties. The reward improvements correlate strongly with physical task success rates, suggesting our \tech effectively captures meaningful progress in manipulation capabilities.

\noindent\textbf{Rollout Entropy of the Policy.}
Appropriate action entropy is crucial: too low restricts exploration, while too high results in unstable improvements. Our implementation maintains moderate entropy levels that gradually decrease as training progresses while allowing initial exploration.

\noindent\textbf{Timing Analysis.}
The time cost distribution for various components of our system is illustrated in the right subfigure of \cref{fig:training_dynamics}. 
With the implementation of GPU-balanced vectorized environments and vLLM acceleration, the time spent on environmental evolution and model rollout has been significantly reduced.
As a result, the primary focus now shifts to the \emph{training phase}, which indicates that further improvements in training efficiency could lead to substantial gains in overall performance.

\subsection{Ablation Study}
\label{subsec:ablation_study}

We further conduct additional ablation experiments on LIBERO-Spatial task suite to more thoroughly validate the proposed design choices in \method, encompassing both the technical aspects and implementation details. The results of these comparative studies are presented in \cref{table:ablation}.
\begin{wraptable}{l}{0.42\linewidth}
\vspace{-0.10in}
\scriptsize
\setlength{\tabcolsep}{8pt}
\renewcommand{\arraystretch}{1.05}
\centering
\label{table:ablation_study} 
\caption{\small \textbf{Ablation Study.} We show the final average success rates on LIBERO-Spatial. Eliminating any individual stabilizing technique from \method leads to rapid collapse, highlighting the critical importance of each technique introduced in \method.
}
\begin{tabular}{l|c}
\toprule
Hyperparameter & LIBERO-Spatial \\
\midrule
\textbf{\method} & \textbf{90.2} \\  
Remove Robotics PRM & 85.8 \\ 
Remove Curriculum & 88.0 \\
$\text{Temperature}=1.5 \!\rightarrow\! 1.0$ & 85.8 \\ 
$\text{Critic Warmup Steps}=5 \!\rightarrow\! 0$ & 80.0 \\ 
$\text{LR}=2e^{-5} \!\rightarrow\! 2e^{-4}$ & 0.2 \\ 
\bottomrule 
\end{tabular}
\label{table:ablation}
\vspace{-0.15in}
\end{wraptable}
\noindent\textbf{Choice of Reward Densification.}
The proposed \tech enhances the success rate from $85.8\%$ to $90.2\%$ compared to the sparse reward baseline, which demonstrates the significance of reward densification. \tech provides more frequent and informative signals that guide the agent toward successful task completion, especially in long-horizon tasks where sparse rewards can lead to inefficient exploration.

\noindent\textbf{Choice of Curriculum Selection Strategy.}
Curriculum-based task selection outperforms uniform random selection, improving success rate from $88.0\%$ to $90.2\%$. By gradually increasing task complexity based on agent performance, the policy learns more effectively on simpler tasks before tackling difficult ones. This reduces catastrophic forgetting and enables better generalization across the task distribution.

\noindent\textbf{Choice of Sampling Temperature.} 
Lower temperatures hamper exploration, causing convergence to suboptimal policies.
To verify this, lowering temperature from $1.5$ to $1.0$ decreases success rate from $90.2\%$ to $85.8\%$, indicating reduced randomness harms policy evaluation and optimization. 

\noindent\textbf{Choice of Critic Warmup.}
Warming up the value model before policy updates improves SR from $80.0\%$ to $90.2\%$. This pre-training provides more accurate feedback for policy gradients, whereas without warmup, initially noisy value estimates negatively impact policy learning.

\noindent\textbf{Choice of Learning Rate.}
A low learning rate ($2e^{-5}$) reaches impressive performance. Higher rates ($2e^{-4}$) cause instability, aligning with the observations in \cite{hu2024flare}. However, low learning rates also slow down convergence, which indicates the importance of implementation details in \ac{rl} systems.

The ablation study highlights the importance of each component in \method, with the full method achieving the highest performance at $90.2\%$ success rate. Removing any single component leads to a noticeable performance drop, demonstrating that all proposed techniques contribute synergistically to the overall effectiveness of our approach.

\subsection{RL vs. SFT}\label{subsec:coverage}

\begin{wrapfigure}{l}{0.5\textwidth}
    \centering
    \vspace{-0.3cm}
    \includegraphics[width=0.45\textwidth]{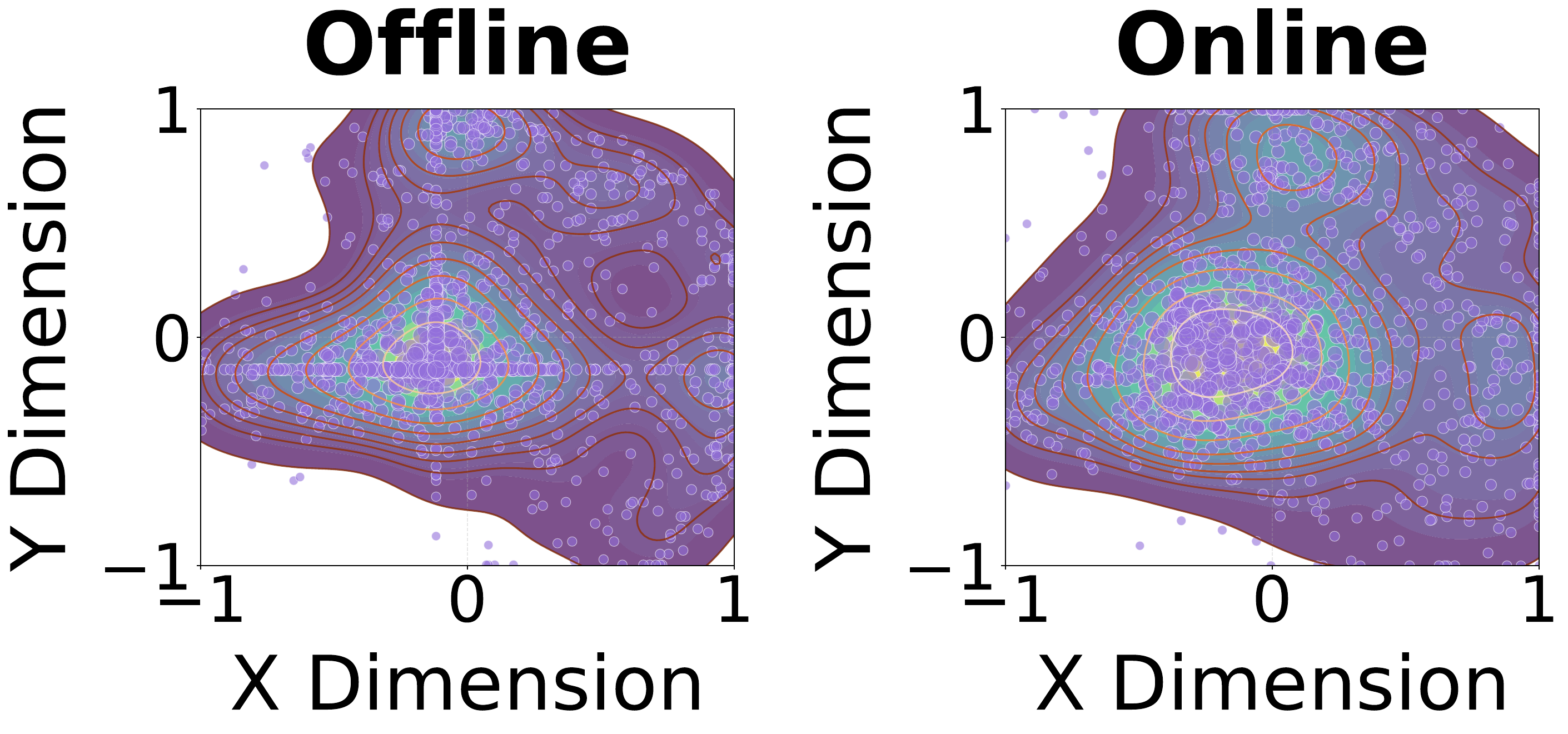}
    \caption{\small \textbf{Action Coverages of SFT and RL methods.} We visualize the first two dimensions of the collected actions (relative end-effector poses) on LIBERO-Spatial.
    }
    \label{fig:state_coverage}
    \vspace{-0.3cm}
\end{wrapfigure}

\noindent\textbf{Action Coverage Analysis.} 
We visualize the projected XY plane of the collected actions in \cref{fig:state_coverage}.
In LIBERO-Spatial, the robot arm is required to navigate to precisely reach a designated contact point. Analysis of the action distributions reveals a distinct contrast between offline and online data collection. Expert actions tend to cluster near the center of the action space, reflecting a preference for positions and often exhibiting repeated motion patterns.
In contrast, actions generated by \ac{rl} agents are distributed more uniformly across the entire action space, which enables RL policies to adapt to a wider variety of states. As a result, the policy trained on RL-collected data exhibits stronger robustness than the SFT model trained on offline data.

\noindent\textbf{Case Study.}
To explore the scenarios where reinforcement learning showcases especial effectiveness, 
\begin{wrapfigure}{l}{0.5\textwidth}
    \centering
    \vspace{-0.5cm}
    \includegraphics[width=0.45\textwidth]{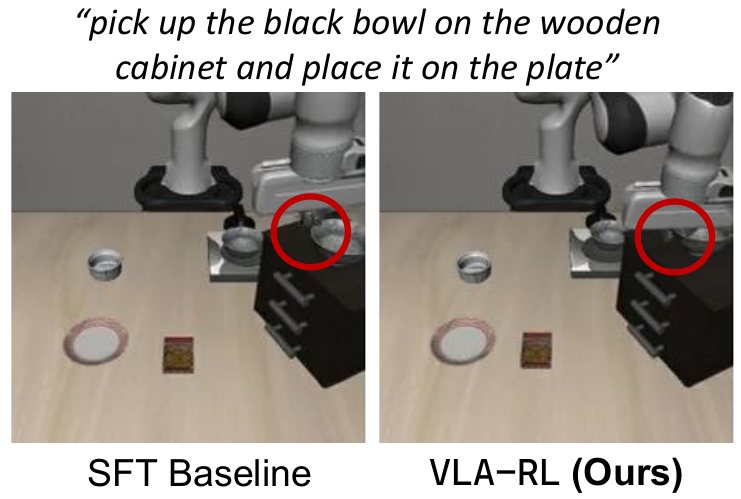}
    \caption{\small \textbf{Case Study.} We show the robot trajectories generated by the starting SFT baseline and \method.
    }
    \label{fig:case_study}
    \vspace{-0.6cm}
\end{wrapfigure}
we analyze the rollout trajectories of both RL and SFT models in LIBERO-Goal.
As shown in \cref{fig:case_study}, the agent is instructed to "pick up the black bowl on the wooden cabinet and place it on the plate", which requires fine-grained interaction. %
The model fine-tuned with \method grasps the bottle successfully, while the SFT baseline fails by trying to grasp at a deviated point. 
We conclude that policies trained with RL-collected data consistently help with alignment issues in contact-rich tasks and reduce premature gripper closure during grasping, aligning with the observations in \cite{xu2024rldg}.

\section{Conclusions and Limitations}
\label{sec:conclusions}

\noindent\textbf{Conclusions.}
We present \method, a scalable reinforcement learning framework that enhances pretrained VLAs through online policy optimization. By formulating general robotic manipulation as multi-modal multi-turn conversations, our approach enables VLA models to explore beyond the limitations of offline demonstration data. Our unified robotic process reward model using fine-tuned vision-language models effectively addresses sparse reward challenges in complex tasks. Through several systematic design choices, we achieve substantial performance gains, surpassing a strong imitation learning baseline OpenVLA-7B by $4.5\%$ on LIBERO benchmarks. Importantly, our observation that performance scales with test-time optimization suggests an emerging principle analogous to inference scaling laws in language models.

\noindent\textbf{Limitations.} \label{para:limitations}
While promising, our approach faces several challenges. For instance, the proposed heuristics for extracting pseudo reward labels may not fully capture the nuances of more dexterous manipulation tasks, potentially leading to inefficient policy optimization.
Future work includes training diffusion-based policies \cite{black2024pi_0,octo_2023} with reinforcement learning beyond auto-regressive VLAs, and exploring online self-improvement with large-scale real-world experience.

\bibliographystyle{splncs04}
\bibliography{reference}

\end{document}